\documentclass[sigconf,screen]{acmart}
\usepackage[utf8]{inputenc} 
\usepackage[T1]{fontenc}    
\usepackage{hyperref}       
\usepackage{url}            
\usepackage{booktabs}       
\usepackage{amsfonts}       
\usepackage{nicefrac}       
\usepackage{microtype}      
\usepackage{lipsum}
\usepackage{graphicx}
\usepackage{footnote}
\usepackage[inline]{enumitem}

\usepackage{algorithm}

\usepackage{graphicx}
\usepackage{multirow}
\usepackage{listings,multicol}
\usepackage{algpseudocode}
\usepackage{lipsum}
\usepackage{fancyvrb}
\usepackage{parcolumns}
\usepackage{verbatim}

\usepackage[T1]{fontenc}
\usepackage{beramono}
\usepackage{xcolor}
\usepackage{balance}

\definecolor{dkgreen}{rgb}{0,0.6,0}
\definecolor{gray}{rgb}{0.5,0.5,0.5}
\definecolor{mauve}{rgb}{0.58,0,0.82}

\usepackage{amsmath}

\lstdefinestyle{myScalastyle2}{
  frame=tb,
  float=*,
  language=scala,
  aboveskip=3mm,
  belowskip=3mm,
  showstringspaces=false,
  columns=flexible,
  basicstyle={\small\ttfamily},
  numbers=none,
  numberstyle=\tiny\color{gray},
  keywordstyle=\color{blue},
  commentstyle=\color{dkgreen},
  stringstyle=\color{mauve},
  frame=single,
  breaklines=true,
  breakatwhitespace=true,
  tabsize=3,
}

\lstdefinestyle{myPythonStyle}{
  frame=tb,
  language=python,
  aboveskip=3mm,
  belowskip=3mm,
  showstringspaces=false,
  columns=flexible,
  basicstyle={\small\ttfamily},
  numbers=none,
  numberstyle=\tiny\color{gray},
  keywordstyle=\color{blue},
  commentstyle=\color{dkgreen},
  stringstyle=\color{mauve},
  frame=single,
  breaklines=true,
  breakatwhitespace=true,
  tabsize=3,
}

\title[IntelliCode Compose: Code Generation using Transformer]{IntelliCode Compose: Code Generation using Transformer}

\author{Alexey Svyatkovskiy}
\authornote{Equal contribution.}
\affiliation{%
  \institution{Microsoft}
  \streetaddress{One Microsoft Way}
  \city{Redmond}
  \state{WA}
  \country{USA}
  \postcode{98052}
}
\email{alsvyatk@microsoft.com}

\author{Shao Kun Deng}
\authornotemark[1]
\affiliation{
  \institution{Microsoft}
  \city{Redmond}
  \state{WA}
  \country{USA}  
  \postcode{98052}
}
\email{shade@microsoft.com}

\author{Shengyu Fu}
\affiliation{
  \institution{Microsoft}
  \city{Redmond}
  \state{WA}
  \country{USA}  
  \postcode{98052}
}
\email{shengyfu@microsoft.com}

\author{Neel Sundaresan}
\affiliation{
  \institution{Microsoft}
  \city{Redmond}
  \state{WA}
  \country{USA}  
  \postcode{98052}
}
\email{neels@microsoft.com}

\copyrightyear{2020} 
\acmYear{2020} 
\setcopyright{acmlicensed}\acmConference[ESEC/FSE '20]{Proceedings of the 28th ACM Joint European Software Engineering Conference and Symposium on the Foundations of Software Engineering}{November 8--13, 2020}{Virtual Event, USA}
\acmBooktitle{Proceedings of the 28th ACM Joint European Software Engineering Conference and Symposium on the Foundations of Software Engineering (ESEC/FSE '20), November 8--13, 2020, Virtual Event, USA}
\acmPrice{15.00}
\acmDOI{10.1145/3368089.3417058}
\acmISBN{978-1-4503-7043-1/20/11}
\begin{document}

\begin{abstract}
In software development through integrated development environments (IDEs), code completion is one of the most widely used features. Nevertheless, majority of integrated development environments only support completion of methods and APIs, or arguments. 

In this paper, we introduce IntelliCode Compose – a general-purpose multilingual code completion tool which is capable of predicting sequences of code tokens of arbitrary types, generating up to entire lines of syntactically correct code. It leverages state-of-the-art generative transformer model trained on 1.2 billion lines of source code in Python, C\#, JavaScript and TypeScript programming languages. 
IntelliCode Compose is deployed as a cloud-based web service. It makes use of client-side tree-based caching, efficient parallel implementation of the beam search decoder, and compute graph optimizations to meet edit-time completion suggestion requirements in the Visual Studio Code IDE and Azure Notebook.

Our best model yields an average edit similarity of $86.7\%$ and a perplexity of $1.82$ for Python programming language.
\end{abstract}

%
%


%

\maketitle

\section{Introduction}

Machine learning has shown a great promise towards improving automated software engineering across all stages. Some of the early applications of machine learning of source code include code search~\cite{codeSearch2016, codeSearch2018}, bug detection and localization~\cite{bugLocalization2016}, program synthesis~\cite{miltos2019Sythesis}, code summarization~\cite{alon2018Summarization} and code completion~\cite{bruch2009learning,hindle2012naturalness,nguyen2013statistical,svyatkovskiy2019Pythia,svyatkovskiy2020fast}.

There are numerous code completion systems capable of effectively recommending method and API calls ~\cite{svyatkovskiy2019Pythia,svyatkovskiy2020fast,bruch2009learning,asad2014methodCC}, or finding the correct argument~\cite{zhang2012ArgumentCC, fazzani2019ArgumentCC, hui2016ArgumentCC}. Majority of argument completion systems would, however, only work when the name of the method or API call is already typed in, thus leaving the task of completing the method calls to software developers. 
 
In this paper, we introduce IntelliCode Compose -- a general-purpose code completion framework, capable of generating code sequences of arbitrary token types, including local variables, methods or APIs, arguments, as well as punctuation, language keywords, and delimiters. IntelliCode Compose serves as a universal programming language modeling tool, effectively generating syntactically correct code in multiple programming languages, capable of completing an entire line of code in a couple of key strokes, with a user experience inspired by Gmail Smart Compose~\cite{gmail}. The proposed system is able to learn to infer types of programming language identifiers and long-range code semantics without inputs extracted by means of a static analyzer explicitly passed to the model as features.

The nature of the problem of code sequence completion makes statistical language modeling approach a promising starting point. To predict a whole line of source code tokens given an existing code context $C$ and vocabulary $V$, we train a neural model to generate tokens $\{m_{t}\}\subset V, t=0...N$, conditioned on a sequence of preceding tokens $\{c_{t}\}, t=0...T$ of code snippet $C$.

The main contributions of the paper are as follows: (i) we introduce and pretrain a multi-layer generative transformer model for code (GPT-C), which is a variant of the GPT-2~\cite{gpt2} trained from scratch on a large unsupervised multilingual source code dataset (cf. sections~\ref{sec:data} and ~\ref{sec:neural}), comparing it to the monolingual counterparts, and a simple n-gram language modeling baseline, (ii) we propose and deploy a novel end-to-end code sequence completion system called IntelliCode Compose based on the GPT-C and an efficient client-side caching system (cf. sections~\ref{sec:decoding} and~\ref{sec:client-postprocess}), (iii) we evaluate the quality of language model pretraining of GPT-C using perplexity, showing that our best model achieves a perplexity of $1.82$; we also show that IntelliCode Compose achieves an average edit similarity of $86.7\%$ (cf. section~\ref{sec:evaluation}), (iv) we introduce MultiGPT-C -- a multilingual version of our model, discuss and compare various approaches to multilingual modeling (cf section~\ref{sec:multilingual}), (v) finally, we discuss and document practical challenges of training intermediate-sized neural transformer models on high-performance computing clusters, and cloud-based model deployment (cf. section~\ref{sec:deploy}).

\section{Motivating Example}

\begin{figure*}
    \includegraphics[width=.49\textwidth]{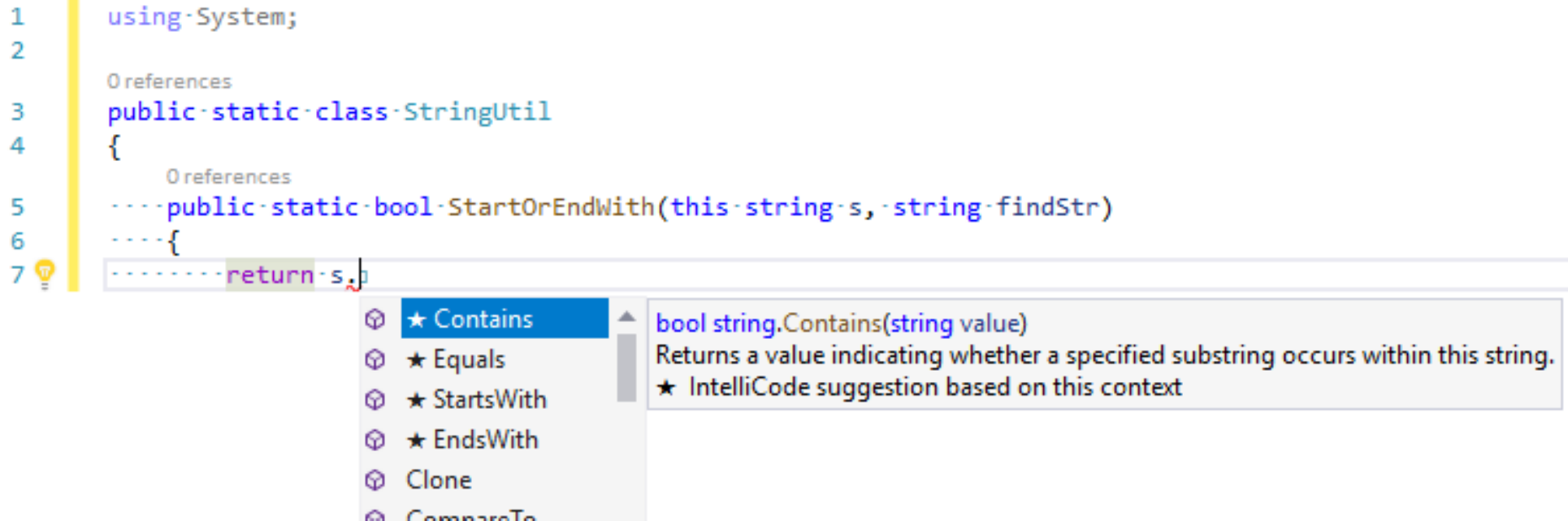}
    \includegraphics[width=.49\textwidth]{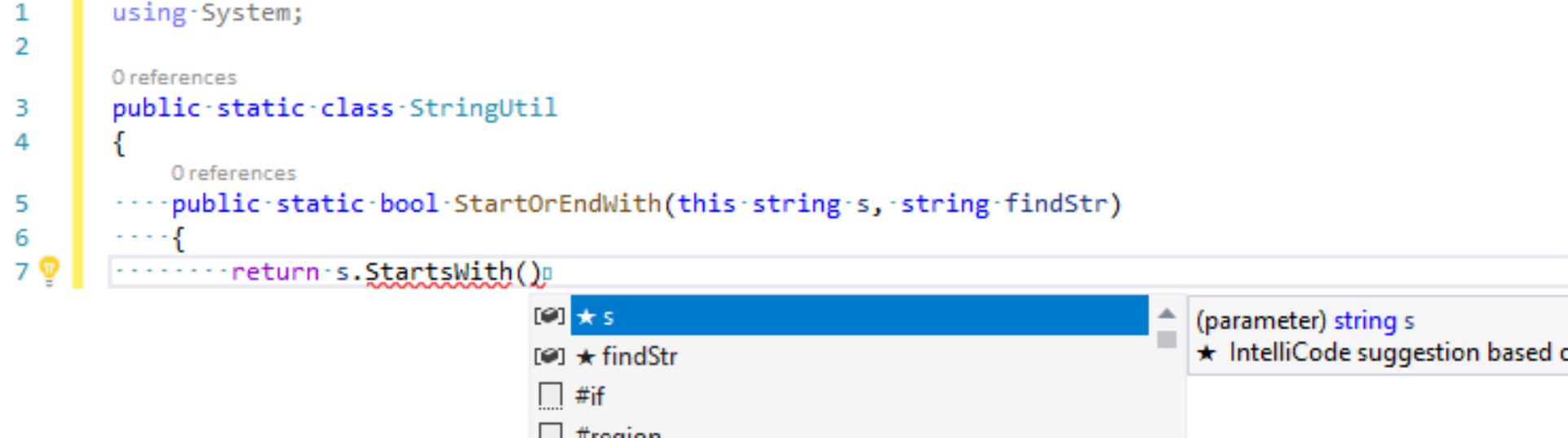}
    \includegraphics[width=.49\textwidth]{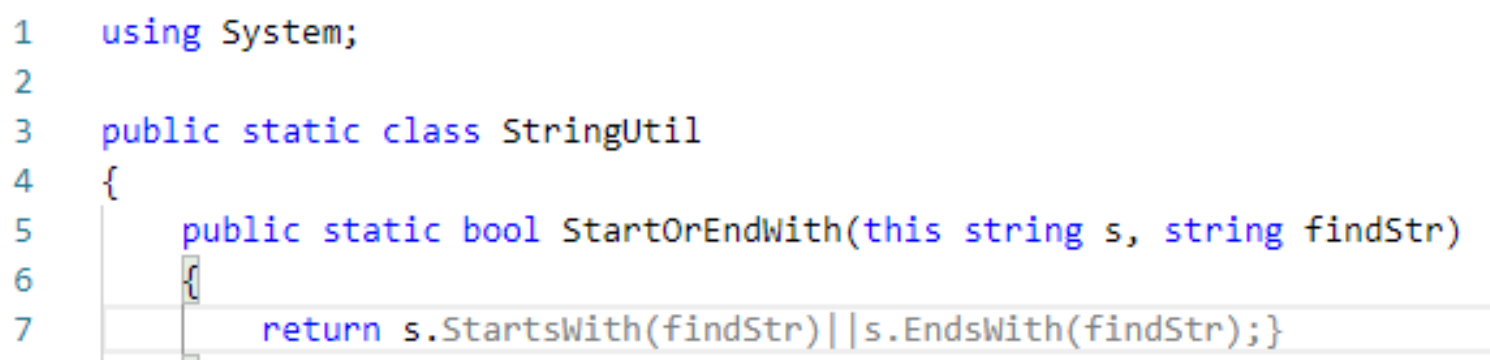}
    \caption{Comparison of code completion scenarios. Top: method completion and argument completion served by \textit{Intellicode}. Bottom: whole-line of code completion served by the \textit{IntelliCode Compose}.} 
    \label{fig:example_completion}
\end{figure*}

Fig.~\ref{fig:example_completion} shows an example method completion and an argument completion in C\# programming language served by the \textit{Intellicode}~\cite{intellicode} extension in Visual Studio IDE\footnote{https://visualstudio.microsoft.com/vs/}, as well as the whole-line of code completion generated by IntelliCode Compose, with the novel completion user experience.
Previously existing code completion tools have focused on specific token types or features, often failing to have a holistic view of the surrounding context. For example, having selected a method to call on the \texttt{s} variable, there are still numerous combinations of arguments to be passed to \texttt{StartsWith}, making this task non-trivial. Correctly suggesting a whole-line of code requires the model to infer types of the target token for method completion, and the correct local variables to be passed as arguments to the methods. Furthermore, additional structural and semantic information needs to be extracted from the context in order to make accurate statement-level suggestions.


\section{Dataset}
\label{sec:data}

We collect a large unsupervised source code dataset to train and evaluate the code sequence completion model. It comprises over 1.2 billion lines of source code in Python, C\#, JavaScript and TypeScript programming languages, as summarized in Tab.~\ref{tab:datasets}. A total of over 52000 top-starred (non-fork) projects in GitHub has been selected, containing libraries from a diverse set of domains, with over 4.7 million source code files. 

We split the dataset into development and test sets in the proportion 70-30 on the repository level. The development set is then split at random into training and validation sets in the proportion 80-20. The final deployed model is retrained using the entire dataset.
\begin{table*}
\caption{Summary of the training dataset.}
\begin{tabular}{llllllllllll} \toprule
\textbf{Programming language} & \textbf{Number of files ($\times10^{3}$)} & \textbf{Number of lines ($\times10^{6}$)} & \textbf{Number of repositories} \\ \midrule
C\# & 1172 & 201 & 4836 \\
Python & 1200 & 240 & 18174 \\
JavaScript & 1982 & 681 &  26553 \\
TypeScript & 437 & 85 & 3255 \\
\bottomrule
\end{tabular}
\label{tab:datasets}
\end{table*}

\section{Approach}
\label{sec:neural}

\subsection{Baseline Code Sequence Completion Model}

We use statistical language modeling approach based on the n-gram model as a baseline in this work. The n-gram model is the probabilistic Markov chain model for predicting text given the context consisting of n-1 preceding tokens.

Given context tokens $\{c_{t}\}, t=0...n-1$, the model estimates the next token probabilities based on the relative frequency counts:
\begin{equation}
   P(m|c_{0}, c_{1}, ... c_{n-1}) = \frac{N(c_{0}, c_{1}, ... c_{n-1}, m)}{\sum{N(c_{0}, c_{1}, ... c_{n-1})}},
\end{equation}
here, numerator gives the count of all $c_{0}, c_{1}, ... c_{n-1}, m$ n-grams in the training corpus, while the denominator gives the cumulative n-gram count that share common prefix $c_{0}, c_{1}, ... c_{n-1}$. During training, the n-grams are extracted by rolling the window of size n sub-tokens, with stride one (see more details on tokenization in section (\ref{seq:tokenize})).

\subsection{Neural Code Sequence Completion Model}
Transformers ~\cite{vaswani2017attention, gpt, bert, xlnet} are a family of neural networks designed to process ordered sequential data. They have found numerous applications in the fields of natural language processing (NLP) and natural language understanding (NLU), including machine translation, question answering, and document summarization. 

Several transformer models such as GPT-2, BERT, XLNet, and RoBERTa~\cite{gpt2, bert, xlnet, roberta} have demonstrated the ability to learn effectively from unlabeled data to perform a wide variety of downstream tasks given supervised discriminative fine-tuning on each specific task. In this work we build on the progress of transformers in NLP and NLU, applying it to an emerging field of \textit{source code understanding}: a form of NLU with additional structural constraints and insights from lexemes, abstract syntax tree (AST) or concrete syntax tree (CST), and dataflow graph.

A transformer block will typically consist of a multi-head self-attention, followed by a two-layer multi-layer perceptron (MLP)~\cite{vaswani2017attention}, optionally containing residual connections and layer normalization~\cite{ba2016layer}. Recent neural architecture searches of transformer models have shown that using depth-wise separable convolutions along with self-attention may speed up training without loss of accuracy~\cite{evolvedTransformer}. A typical transformer architecture for a sequence-to-sequence task will have an encoder (a stack of transformer blocks) and a decoder stack. Unlike the vanilla recurrent neural networks (RNNs) or their gated variants, including LSTM and GRU, transformers do not require tokens in a sequence to be processed in a specific order, thus allowing more options for training parallelization~\cite{shoeybi2019megatronlm}. Composed of feed-forward layers, convolutions, and self-attention, transformers are easy to quantize and serve in production.

GPT-2 is an auto-regressive pre-trained model consisting of a decoder-only transformer stack and one or more output layers, often referred to as ``heads''. GPT-2 for language modeling task has a linear output layer with \texttt{softmax} output activation. IntelliCode Compose is built around a multi-layer generative pretrained transformer model for code (GPT-C), which is a variant of the GPT-2 trained from scratch on source code data, with weights of the output linear layer tied to the input embedding matrix, having specific hyperparameters as described in Tab.~\ref{tab:example_hyperparameters}. 


In order to predict a sequence of response tokens $M = \{m_{t}\}, t=0...N$, conditioned on code snippet typed in by a software developer $\{c_{t}\}, t=0...T$, we need to estimate the following conditional probability distribution:
\begin{eqnarray}
P(m_{0}, m_{1}, ... m_{N}|c_{0}, ... c_{T}) = \prod_{i=1}^{N} P(m_{i}|c_{0}, c_{1}, ... c_{T}, m_{0}, ... m_{i-1}).
\end{eqnarray}

With the autoregressive approach, the objective is to maximize the following log-likelihood:
\begin{eqnarray}
    L(M) = \sum_{i}\log{P(m_{i}|c_{0}, c_{1}, ... c_{T}, m_{i-k}, m_{i-k+1}, ... m_{i-1}; \Theta)}
\end{eqnarray}
where $k$ is the length of predicted code sequence, and the conditional probability $P$ is modeled using a neural network with parameters $\Theta$. These parameters are learned via stochastic gradient descent optimization procedure.

GPT-C applies a multi-headed self-attention operation over the input context tokens followed by position-wise feed-forward layers to produce an output distribution over target tokens:
\begin{eqnarray}
\label{eq:forward_eqns}
h_0 &=& W_e\cdot C + W_p, \\
h_l &=& \texttt{transformer\_block}(h_{l-1}), \forall l = 1...n,  \\
P(m_{t}) &=& y_t = \texttt{softmax}(h_n\cdot W_e^T ), t=0...N, \\
\end{eqnarray}
where $C = c_{-k}, c_{-k+1}, ... c_{-1}$ is the context vector of tokens, $n$ is the number of layers, $W_e \in R^{|V| \times d_{x}}$ is the tokens embedding matrix, and $W_p \in R^{N_{ctx} \times d_{x}}$ is the position embedding matrix, which encodes relative positions of tokens in a sequence. $N_{ctx}$ is the length of the sequence attended to (context length), $|V|$ is vocabulary size, and $d_{x}$ is the embedding dimension. 

We are reusing the input token embedding matrix as the output classification matrix~\cite{socher16Tieweights}, which allows to remove the large fully connected layer reducing the number of parameters by 25\%. More specifically, we introduce a projection matrix $A = (a)_{ij} \in R^{d_{model}\times d_{x}}$ initialized according to a random uniform distribution. Given an encoded code context and a hidden state at the last temporal step $h_n(T) \in R^{d_{model}}$, we obtain the predicted token embedding vector by multiplying the two together as $W_{e}^{pred} = (\textit{w}^{pred})_{j} \in R^{d_{x}}$ as:
\begin{equation}
\textit{w}^{pred}_{j} = \sum_{i} h_{ni}(T)a_{ij}.
\end{equation}
Subsequently, the logits are obtained as:
\begin{equation}
y_k = \sum_{j} \textit{w}_{kj}\textit{w}^{pred}_{j} + b_{k} 
\end{equation}
where ${b}_{k}, k = 0...|V|-1$ is the bias vector, and $d_{model}$ is the number of hidden units per transformer block.

During inference, beam-search decoding algorithm is applied to iteratively extract best token sequences according to a negative log-likelihood optimization objective. This is explained in more detail in section~\ref{sec:decoding}.

\section{Preprocessing}
\label{sec:preprocessing}

In what follows, we treat the source code data as a sequence of tokens corresponding to the output of a lexical analyzer. Incidentally, this can also be constructed through an in-order traversal of the terminal nodes of a concrete syntax tree (CST). In this work, we do not leverage high-level structural representation such as abstract or concrete syntax trees or control flow graphs, as it introduces additional overhead and dependencies which slows down the inference and reduces coverage of the code completion system. Additionally, for most programming languages, such representations can only be correctly retrieved on complete code snippets that are syntactically correct, which is often not available for a code completion system.

Our approach is based on statistical language modeling of source code, with several normalization rules extracted from concrete syntax tree of a program. 
To overcome the issue of different styles and white space or tab conventions, we transform the code into symbolic program tokens using custom tokenizers and regenerate the code with a common style. During preprocessing, we parse program code in each file, extract information about token types and apply it to normalize the code, extract subtoken vocabulary and encode the sequences. This is done both for training and inference.

\subsection{Overcoming a Closed Vocabulary Problem}
\label{seq:tokenize}
 
A typical language model will attempt to generate a probability distribution over all tokens in the vocabulary. This requires the model to have access to encodings of all such tokens. In vanilla language models this is achieved with a fixed vocabulary matrix, thus limiting model coverage to unseen tokens.
 
The issues of coverage can be addressed by using finer-level encodings for tokens. Instead of learning representations for each token, we learn representations for subtokens or combinations of Unicode characters. This both reduces the need to store an entire vocabulary and makes the model more robust to out-of-vocabulary tokens. This allows us to potentially generalize to previously unseen methods, APIs, other language identifiers, or even training code completion models for multiple programming languages. 

We experiment with two specific ways of tokenization:
\begin{enumerate}
    \item Byte-Pair Encoding (BPE) tokenization -- unsupervised tokenization, in which the most frequently occurring pair of Unicode characters is recursively replaced with a character that does not occur in the vocabulary -- the approach adopted by various contextual language models in NLP.
    \item Tokenization by splitting programming language identifiers using casing conventions, such as \texttt{camelCase}, and\\ \texttt{PascalCase} or \texttt{snake\_case} -- the approach that has been shown to work for programming languages, though not applicable to natural languages.
\end{enumerate}

\begin{figure*}
\begin{center}
    \includegraphics[width=.80\textwidth]{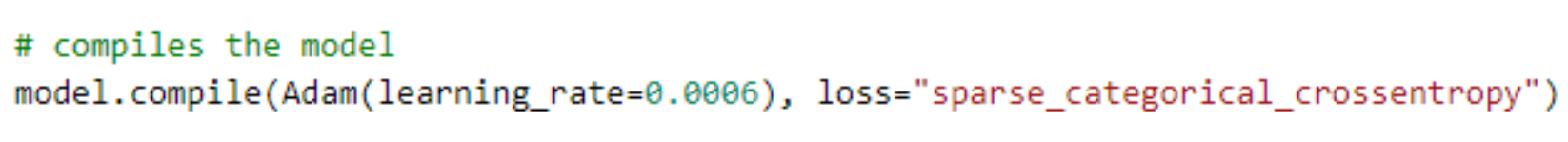}
    \includegraphics[width=.80\textwidth]{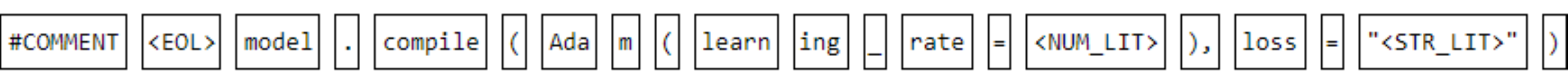}
    \includegraphics[width=.80\textwidth]{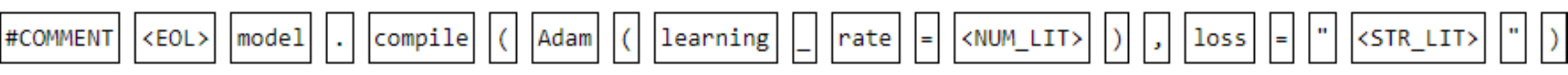}
\caption{Illustration of tokenization approaches. From top to bottom: a raw code snippet consisting of a comment and an API call with arguments, corresponding BPE split using \textit{sentencepiece}, and corresponding split based on casing conventions.}
\label{fig:tokenize}
\end{center}
\end{figure*}

We use the \textit{sentencepiece}\footnote{https://github.com/google/sentencepiece} tokenizer to extract subtoken level vocabulary, with special tokens for control flow and code structure representation. More specifically, we add control flow tokens \texttt{<BOF>} and \texttt{<EOF>} to mark the beginning and ending of a file in order to disambiguate similar identifier names in different files, and \texttt{<EOL>} to mark the ending of a line. Additionally, since Python uses white-spaces and indentation to demarcate code scope, we introduce \texttt{<INDENT>} and \texttt{<DEDENT>} tokens to represent those scope delimiters. Fig.~\ref{fig:tokenize} illustrates the tokenization approaches.

\subsection{Exposing Sensitive Data through Code Suggestions}
\label{sec:preprocessing-literals}

\begin{figure}
\begin{center}
    \includegraphics[width=.49\textwidth]{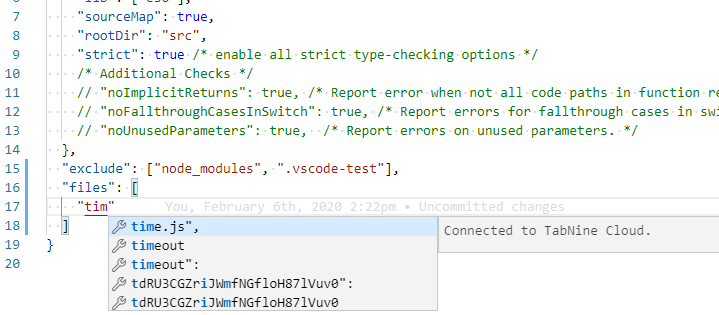}
\caption{Example completion served by the \textit{TabNine} system exposing a fragment of hash value.}
\label{fig:tabning_hash}
\end{center}
\end{figure}

Production-level code completion systems based on statistical language modeling are commonly trained on vast amounts of source code mined from GitHub or other version control systems. As large amount of public data is ingested, it is unavoidable to encounter cases where people unintentionally leave sensitive information in their code, as part of string literals, code comments, or configuration files. Fig.~\ref{fig:tabning_hash} shows an example completion served by the \textit{TabNine}~\cite{tabnine} system exposing irrelevant and potentially sensitive data. 
To tackle this problem, the training process needs to be shielded from inadvertently gaining access to secrets or personally identifiable data. For this reason, we identify and normalize numeric literals, string literals and comments, including docstrings, to \texttt{<NUM\_LIT>}, \texttt{<STR\_LIT>} and \texttt{<COMMENT>} special tokens\footnote{For C\#, in addition to \texttt{<STR\_LIT>} and \texttt{<NUM\_LIT>}, we also introduce \texttt{<CHAR\_LIT>} for character literals. For JavaScript, we also introduce \texttt{<RE\_LIT>} for regular expression literals.}, respectively. 
However, we have found that the most frequently used literals often contain relevant information and can be used directly in the completions. For each language, a number of top most frequent numeric and string literals are preserved as \texttt{<STR\_LIT:lit>} where \texttt{lit} is the original literal. For instance: \texttt{"\_\_main\_\_"}, \texttt{"POST"}, \texttt{"en"}, \texttt{"default"}.  We did, however, leave identifier names to make code suggestions context-dependent. The exact number of literals kept as well as the percentage they represent in the training data are shown in Table.~\ref{tab:literals}.

\begin{table}
    \caption{Number of literals kept for each type, and the percentile they represent for each training dataset.}
    \centering
    \begin{tabular}{lllll} \toprule
         & Number kept & Python & C\# & JS,TS \\ \midrule
        String & 200 & 18 & 13 & 20 \\
        Number & 50 & 63 & 58 & 70 \\
        Char & 20 & - & 42 & - \\
        RegEx & 50 & - & - & 17 \\
        \bottomrule
    \end{tabular}
    \label{tab:literals}
\end{table}

\section{Model Training}

Optimizing transformer neural networks is a computationally intensive problem which requires the engagement of high-performance computing (HPC) clusters in order to improve time to solution. Selection of well-performing hyperparameters requires searching a high-dimensional space. To evaluate a neural architecture or a set of hyperparameters entails running full model training and inference.

We scale up the training using synchronous data-parallel distributed training algorithm with local gradient accumulation.
The learning rate controlling the magnitude of the weight update during gradient optimization is lowered upon completion of each epoch according to the cosine decay. In a distributed regime, we increase the learning rate during the first few epochs (``warm-up'' period) to facilitate reliable model convergence. 

The offline training module of the IntelliCode Compose system is implemented as a Python library integrating PyTorch and Horovod~\cite{sergeev2018horovod} with Adasum algorithm for gradient summation\footnote{https://github.com/horovod/horovod/pull/1484}. The software stack makes use of CUDA 10, GPU accelerated deep learning primitives from CuDNN 7, and PyTorch 1.2.0, NCCL collective communication library. 
We have trained our models on 5 Lambda V100 boxes, each having sixteen V100 GPUs with $32~$GB HBM2 memory, eight $100~$GB InfiniBand, and one $100~$GB Ethernet connection, managed with Kubernetes. 

With the data-parallel implementation, pure computation time $T_{batch}$ per mini-batch step remains constant in the number of worker GPUs. The amount of data processed during one mini-batch step increases linearly with the number of engaged workers $N$. Synchronization between workers performed by means of a tree-like \texttt{allreduce}, would yield logarithmic complexity $T_{sync} \propto \log{N}$. Thus, the number of mini-batches would decrease linearly with $N$, giving a following scaling model:
\begin{equation}
  T_{epoch} = \frac{1}{N}\cdot(T_{batch} + T_{sync}) = \frac{1}{N}\cdot(A + B \cdot log(N)) = O(\frac{log(N)}{N})   
\end{equation}

\begin{figure}
\begin{center}
    \includegraphics[width=.47\textwidth]{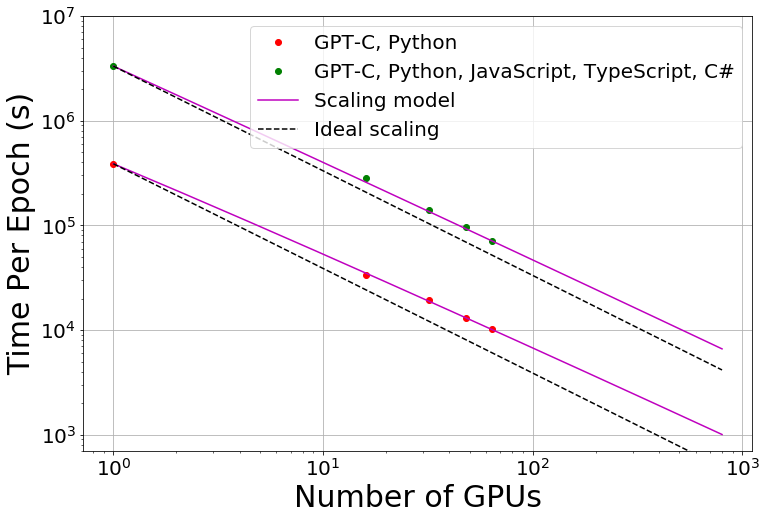}
    \includegraphics[width=.47\textwidth]{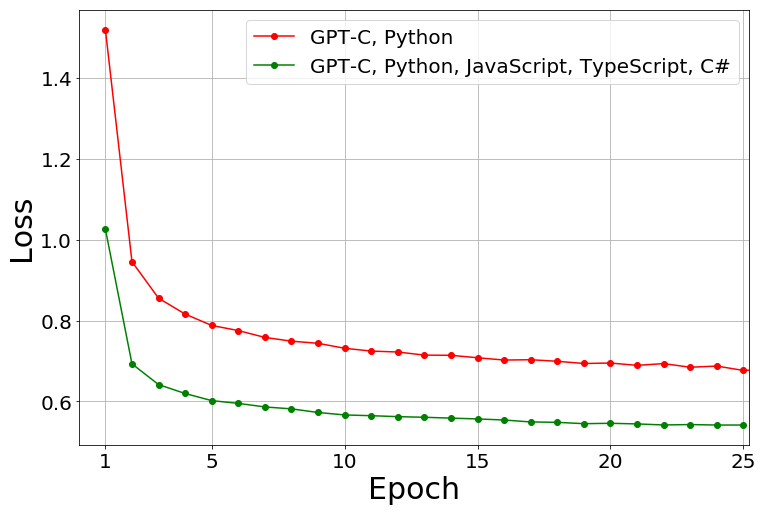}
    \caption{Top: time required to complete one pass over the dataset (one "epoch") during training versus the number of worker GPUs engaged. Experimental data are compared with a semi-empirical theoretical scaling model and ideal scaling. Bottom: training loss as a function of epoch for monolingual and multilingual models.} 
\end{center}
\end{figure}

\begin{table}
\caption{Neural network training performance summary for monolingual (Python) and multilingual (C\#, Python, JavaScript and TypeScript) 24-layer GPT-C models on 80 GPU workers.}
\begin{center}
\begin{tabular}{llllllllllll} \toprule
 & Monolingual & Multilingual \\ \midrule
\textbf{Cumulative batch size} & 160 & 160 \\
\textbf{Time per epoch}& 2.8 hours & 19.7 hours\\
\textbf{Samples / second} & 163 $\pm$ 5 & 148 $\pm$ 5 \\
\textbf{Tokens / second} & 167000 $\pm$ 5000 & 152000 $\pm$ 5000 \\
\bottomrule
\end{tabular}
\label{tab:trainingtimes}
\end{center}
\end{table}

Overall, the model architecture, tokenization, and training procedure produce a large number of hyperparameters that must be tuned to maximize predictive performance. These hyperparameters include numerical values such as the learning rate and number of transformer layers, dimension of embedding space, but also abstract categorical variables such as the precise model architecture or the source code normalization algorithm. The number of trainable parameters in the GPT-C transformer model scales near-linearly as a function of number of transformer blocks, and quadratically with the number of hidden units per block as: $d_{x} \cdot (|V| + N_{ctx}) + A\cdot n \cdot d_{model}^2$. The constant $A$ here is defined by the parameters of the MLP part of the transformer.  

The best performing monolingual GPT-C models have 24 layers, scaled dot-product attention with 16 heads, and are trained with BPE vocabulary size of 50000, while the best multilingual version has 26 transformer layers, 16 heads, and the vocabulary size of 60000 subtokens. The rest of the model architecture parameters is summarized in Tab.~\ref{tab:example_hyperparameters}.

\begin{table}
\caption{Well-performing values of model architecture hyperparameters.}
\begin{center}
\begin{tabular}{llllllllllll} \toprule
\textbf{Hyperparameter} & \textbf{Explanation} & \textbf{Best value} \\ 
\midrule
$d_{model}$ & Hidden units per layer & 1024 \\
$N_{ctx}$ & Code context length & 1024  \\
$d_{x}$ & Embedding dimension & 1024  \\
$N_{head}$ & Attention heads & 16  \\
Dropout & Dropout keep probability & 0.9 \\
\bottomrule
\end{tabular}
\label{tab:example_hyperparameters}
\end{center}
\end{table}
We train GPT-C using Adam stochastic optimization scheme with weight decay fix, base learning rate of $6.25\times10^{-5}$, cumulative batch size of 128, learning rate decay of 0.98 per epoch, and categorical cross-entropy loss. For multilingual model, each training mini-batch has to be composed of sentences coming from the same language, which is sampled at random from the set of all available languages.

\section{Sequence Decoding}
\label{sec:decoding}

Each inference call to the model yields a probability distribution vector over subtokens in the vocabulary. This can be conceptualized as an $N$-ary tree of subtokens rooted in the last subtoken of the code context typed in by a developer. The depth of the tree is defined as a length of the desired completion sequence. Each code sequence suggestion is effectively a path on the tree, from the root node to a terminal node. The beam search algorithm is employed to explore and rank those paths, improving recommendation relevance of code sequences. At every step, the results are aggregated and the top \textit{k} results are selected, where \textit{k} is the beam width. Decoding continues for a preset number of subtokens or until a break token is reached. The set of break tokens includes the \texttt{<EOL>} (end-of-line) token as well as other language-specific tokens that often precede end-of-line under common code style patterns. 

A naive beam search implementation would iterate over the top \textit{k} candidates at every step to produce the output vector. However, for a sequence of length $L$, this would require $L\times k$ inference calls to the model, significantly increasing the inference time and degrading the overall real-time user experience. Instead, we aggregate top \textit{k} candidates and perform batched inference calls at every decoding step, which reduces the number of inference calls to $L$. Tab.~\ref{tab:beamsearch-comparison} provides a comparison of the inference speeds for scenarios with different beam widths and sequence lengths, quoting speed-ups gained through parallelization.

Given sequential nature of the beam search decoding, we cache the attention keys and values of the transformer blocks as computed by the model for previous step (token), passing it as input to the model at the current step instead of recalculating from scratch. This further speeds up inference by 10\%. The speed improvement with parallel and cached search is most apparent for large $L$.

\begin{table}
\caption{Inference speed comparison for search scenarios with different beam widths $k$ and sequence lengths $L$, with different beam search setup.}
\begin{center}
\begin{tabular}{llllllllllll} \toprule
$L$ & \textit{k} & \textbf{Sequential (ms)} & \textbf{Parallel (ms)} & \textbf{Cached (ms)} \\ \midrule
10 & 1 & 250 & 250 & 220 \\
10 & 10 & 1700 & 1000 & 820 \\
25 & 15 & 7500 & 3000 & 2700 \\
\bottomrule
\end{tabular}
\label{tab:beamsearch-comparison}
\end{center}
\end{table}

\section{Client-Side Post-Processing}
\label{sec:client-postprocess}

\subsection{Completion Caching}
During our user experience study, we have found that a response time under $100~$ms is necessary to avoid any feeling of delay or lag. To achieve this in a cloud-based model deployment setting, we introduce caching on the client-side. Any time a developer types a non-alphanumeric character, suggestions are queried from the server. Those suggestions, each as a list of tokens along with their scores, are stored into a trie\footnote{A trie is a tree-like data structure where each node is a sub-string and strings can be composed by traversing down a path from the root.}, and this trie is then placed into a cache. The cache key is the piece of code preceding the point where the suggestion was queried. This approach allows us to prune the tree efficiently at a character-level as the user continues typing. To obtain the final completion suggestion, we simply traverse this tree greedily by always branching to the node with the highest score. 

\begin{figure*}
    \includegraphics[width=.80\textwidth]{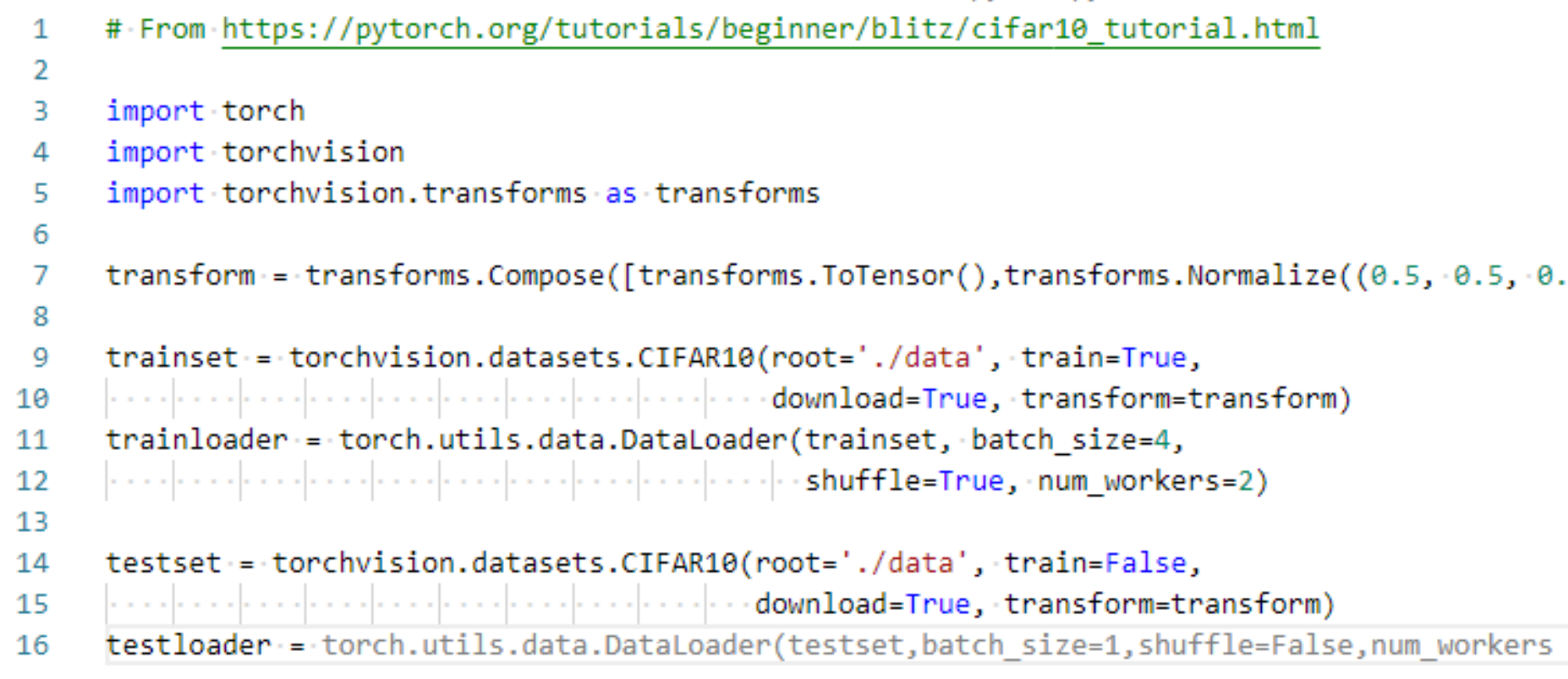}
    \includegraphics[width=.95\textwidth]{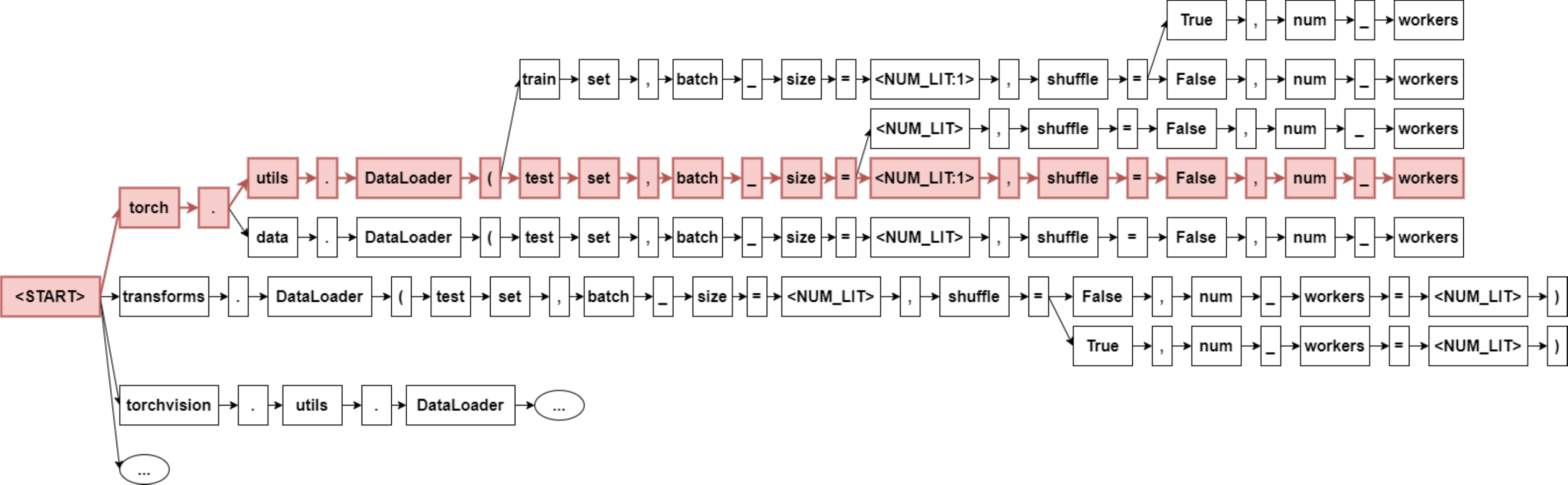}
    \caption{Top: code snippet and the completion suggestion served by the model in Visual Studio Code. Bottom: completion-tree reconstructed by the client application, with the highlighted path shown in red (lower section of the tree is truncated to reduce visual clutter).} 
    \label{fig:example_completion3}
\end{figure*}

Through experimentation, we have found that the model occasionally returns multiple similar but equally valid suggestions. In order to preserve accuracy, we terminate the completion-tree traversal if none of the child nodes has a score that is equal to or larger than the score of its parent multiplied by a ratio $R$, defined as:
\begin{equation}
    R = \frac{\alpha}{1+e^{\frac{-L}{\kappa}}}.
\end{equation}
This early-stopping allows us to break the suggestion at points where the model is equally confident in multiple valid suggestions. Here, $L$ is the position of the root node of the trie, $\alpha$ is the relaxation factor, and $\kappa$ is the curvature factor. $\alpha$ is used to adjust the values of $R$ for very small or very large values of $L$. A lower value of $\alpha$ would relax the policy producing longer completion suggestions, while a value closer to $1.0$ would tighten the policy producing shorter suggestions. $\kappa$ controls the rate of increase of the $R$. A smaller $\kappa$ would give a steeper curve for smaller values of $L$, producing shorter suggestions, while a larger value of $\kappa$ would yield a flatter curve resulting in longer completion suggestion. In our deployment, we select $\alpha=0.8$ and $\kappa=10$ to gain a balance between suggestion length and relevance. We have found this approach to yield the highest increase in productivity, allowing software developers to retain a certain level of control over the suggestions. Fig.~\ref{fig:example_completion3} shows an example code completion suggestion and the corresponding completion-tree.

\subsection{Suggestion Processing}
As mentioned in section~\ref{sec:preprocessing}, we introduce several new tokens that are not present literally in the input code. As we decode the output sequences, the model would generate those new tokens at the appropriate locations. In order to incorporate those tokens fluently into the completion sequences, we need to post-process them on the client side into printable characters using the following rules:
\begin{enumerate}
    \item {\texttt{<BOF>} and \texttt{<EOF>} tokens are ignored, as they almost never seen in the suggestion sequence and do not provide any additional information relevant to the completion sequence.}
    \item {\texttt{<EOL>} serves as a break token for beam search decoder. We truncate the completion at this token as it indicates the end of the line.}
    \item {\texttt{<STR\_LIT>} and \texttt{<NUM\_LIT>} tokens become placeholders and are replaced by the default literals (empty string and number 0, respectively). Visual Studio Code provides the ability to insert code snippets with placeholders in them, which the user can easily navigate through using \textbf{TAB} key. For raw literal tokens such as \texttt{<STR\_LIT:\_\_main\_\_>}, as mentioned in section~\ref{sec:preprocessing-literals}, we use the raw literal image as the placeholder instead.}
\end{enumerate}

\section{Multilingual Model}
\label{sec:multilingual}

Multilingual approach allows low-resource programming languages to benefit from more popular languages in terms of modeling quality. Multilingual models also hold a promise of being easier to maintain and serve in production.

To prepare a multilingual version of the IntelliCode Compose, we have extracted a shared sub-token vocabulary for Python, C\#, JavaScript and TypeScript programming languages using the BPE tokenizer. We explored and compared the following four ways of training multilingual GPT-C models:
\begin{enumerate}
    \item \textit{Language-agnostic baseline} completely disregards the language type information, effectively attempting to train the model as monolingual. We have found this approach to underperform significantly as compared to the monolingual versions for each language.
    \item \textit{Language-type embedding}. Looking for a stronger baseline, we introduced the language type embedding matrix $W_l \in R^{N_{lang} \times d_{x}}$, combining it via addition with the token and position embedding matrices of GPT-C model for each token in the sequences during the forward pass, given by Eq.~\ref{eq:forward_eqns}, according to: $h_0 = W_e \cdot C + W_p + W_l$. $N_{lang}$ denotes the number of programming languages in the training dataset.
    \item \textit{Language specific control codes}, is an approach introduced in~\cite{T5,CTRL} that uses target prefixes to facilitate constrained language modeling or multitask learning. In what follows, for each programming language we insert a sequence of tokens in the beginning of each training sample according to: "$\mathrm{lang}~*$ remaining token sequence", where $\mathrm{lang} \in \{\mathrm{Python}, \mathrm{C\#}, \mathrm{JavaScript}, \mathrm{TypeScript}\}$. In expectation, control codes would signal the neural network that a given sequence belongs to a particular programming language. As shown in Tab.~\ref{tab:multilang_summary} this approach works rather well, yielding results comparable with monolingual counterparts.
    \item Finally, we add \textit{a programming language classification task during model pretraining}, an approach inspired by natural language inference (NLI) RocStories and SWAG tasks~\cite{mostafazadeh-etal-2016-corpus,zellers2018swag}. As such, the model is trained using two optimization objectives: language modeling and multiple choice classification to detect programming language. Given $N_{lang}$ one-dimensional monolingual arrays of subtokens. 
\end{enumerate}


\begin{table*}
\caption{Detailed evaluation results for various multilingual modeling approaches based on GPT-C. Model performance metrics are reported on multilingual test sample.} 
\centering
\begin{tabular}{llllllllllll} \toprule
	\textbf{Model}   &  \textbf{PPL} &
			\multicolumn{2}{c}{\textbf{ROUGE-L}} & \textbf{Edit similarity (\%)} & \textbf{Model size} \\ \cmidrule{3-4} \cmidrule{7-10}
	                & & Precision &  Recall \\ \midrule
Baseline & 2.15 & 0.25 & 0.24 & 56.3 & 374M \\	
Language Embedding & 1.94 & 0.52 & 0.66 & 71.7 & 379M \\
Control codes & 1.73 & 0.64 & 0.75 & 81.5 & 374M \\
MultiGPT-C (double heads)  & 1.65 & 0.66 & 0.76 & 82.1 & 374M \\	
\bottomrule
\end{tabular}
\label{tab:multilang_summary}	
\end{table*}
As seen from Tab.~\ref{tab:multilang_summary}, language specific control codes provide necessary supervision to constrain language generation for each specific language. The multitask approach of language modeling combined with multiple choice classification provides a further improvement. As such, the double heads multilingual model, referred to as MultiGPT-C throughout the paper, is selected for multilingual deployment candidate.

\section{Evaluation}
\label{sec:evaluation}

\subsection{Evaluation Metrics}

We use perplexity to evaluate the quality of language model pretraining of GPT-C models, defined as:
\begin{equation}
    PPL = exp(-\sum_{i}^{T}{P(x_{i})log{P(x_{i})}}), \forall i \in 0...T.
\end{equation}
where $x_{i}$ is the truth label and $P(x_{i})$ is the model output. A model with lower perplexity assigns higher probabilities to the true tokens, and is expected to perform better.

Besides perplexity, we consider two evaluation metrics to measure offline performance of the code sequence completion system: the Recall-Oriented Understudy for Gisting Evaluation score (ROUGE)~\cite{rouge} and the Levenshtein similarity.

The output sequences served by the IntelliCode Compose may consist of up to 20--30 characters, including identifier, literals, language keywords, punctuation marks, whitespaces, and delimiters. From our online telemetry, we have observed that the accepted suggestions are on average 25 characters long, with 25\% of accepted suggestions being longer than 45 characters.
Accuracy could measure correctness of the exact match, failing, however, to capture the proximity when a completion suggestion partially matches the target sequence, which could still be a valid completion suggestion. 

For instance, given model suggestion: \texttt{tf.train.\linebreak AdamOptimizer(learning\_rate=<NUM\_LIT>)} and target sequence \texttt{tf.train.GradientDescentOptimizer()}, we see that the model correctly captures the intent of software developer to create an optimizer, suggesting the \texttt{AdamOptimizer} with the learning rate parameter, while the target is the standard gradient descent optimizer, with default value of the learning rate parameter.

The ROUGE score is the metric commonly used to evaluate machine translation models. Its ROUGE-L variant is based on the Longest Common Subsequence (LCS)~\cite{lcs} statistics. LCS takes into account structure similarity and identifies longest co-occurring $n$-grams.

The Levenshtein distance measures how many single-character edits -- including insertion, substitution, or deletion -- does it take to transform one sequence of tokens to another. Quite often, even if a suggested completion is only an approximate match, developers are willing to accept it, making appropriate edits afterwards. As such, the Levenshtein edit similarity is a critical evaluation metric.


ROUGE score and edit similarity capture string similarity of completion suggestions and the target code. However, not all kinds of imperfections in the code suggestions generated by the model are accepted equally by the end users. In particular, syntax errors are relatively easy-to-find bugs which are considered ``silly'' by software developers. We utilize tree-sitter~\footnote{https://tree-sitter.github.io/tree-sitter/} parser to estimate syntactic correctness of the completion suggestions generated by the tool, by parsing the file-level code context together with the suggestion generated by the model. Given the IntelliCode Compose line-of-code completions may represent partial code statements, for this experiment, we remove end-of-line token from the list of break tokens for beam search decoder, capturing the latest complete statement decoded. We use the Black~\footnote{https://black.readthedocs.io/en/stable/} formatter to compress multi-line statements to single lines to minimize the amount of partial code statements generated by the tool. We observe 93\% of completion suggestions in Python programming language are syntactically correct. By visually inspecting several completion suggestions which have failed to parse by tree-sitter, we conclude that roughly a half of those can be attributed to partially generated statements.


\subsection{Evaluation Results}

We start by comparing our monolingual GPT-C model for Python programming language against the simple n-gram language models with n = 3, 5, and 7. As seen in Tab.~\ref{tab:baseline_table}, performance of the simple n-gram language model is significantly lower, improving for intermediate n, then drops again for n = 7. Inspecting model suggestions visually, we identified that the n-gram approach is highly sensitive to out-of-vocabulary tokens, especially for seven-gram, which represents the main limitation of this approach.
\begin{table}
\caption{Python n-gram language model baseline performance comparison against the monolingual GPT-C model pre-trained on Python programming language.} 
\centering
\begin{tabular}{lllllllllll} \toprule
	\textbf{Model}   &
			\multicolumn{2}{c}{\textbf{ROUGE-L}} & \textbf{Edit similarity (\%)} \\ \cmidrule{2-3} \cmidrule{5-8}
	                & Precision &  Recall \\ \midrule
\textbf{3-gram LM} & 0.16 & 0.28 & 37.8 \\ 
\textbf{5-gram LM} & 0.40 & 0.45 & 59.7 \\ 
\textbf{7-gram LM} & 0.34 & 0.34 & 39.2 \\ 
\textbf{GPT-C} & 0.80 & 0.86 & 86.7 \\	
\bottomrule
\end{tabular}
\label{tab:baseline_table}	
\end{table}

Next, Tab.~\ref{tab:main_results_table} provides a detailed summary of evaluation results for a selected subset of well-performing models. 
\begin{table*}
\caption{Detailed evaluation results for our best-performing monolingual (GPT-C) and multilingual (MultiGPT-C) models, including the zero-shot performance of Python model on C\# programming language, as well as the open source HuggingFace~\cite{Wolf2019HuggingFacesTS} medium-sized GPT-2 checkpoint pretrained on WebText~\cite{Gokaslan2019OpenWeb} dataset. Multilingual model performance metrics are reported separately for each of the programming languages from the training set.} 
\centering
\begin{tabular}{llllllllllll} \toprule
	\textbf{Model}   &  \textbf{Test (Train) Languages}    &  \textbf{PPL} &
			\multicolumn{2}{c}{\textbf{ROUGE-L}} & \textbf{Edit similarity (\%)} & \textbf{Model size} \\ \cmidrule{4-5} \cmidrule{8-11}
	                & & & Precision &  Recall \\ \midrule
\textbf{GPT-C} & \textbf{C\#} (C\#) & 1.91 & 0.57 & 0.79 & 76.8 &  366M\\	
\textbf{GPT-C} & \textbf{Python} (Python) & 1.82 & 0.80 & 0.86 & 86.7 & 366M \\	
\textbf{GPT-C} & \textbf{JS,TS} (JS,TS) & 1.40  & 0.58 & 0.72 & 84.1 & 366M \\	
\textbf{GPT-C}, zero-shot & \textbf{C\#} (Python) & -- & 0.39 & 0.50 & 57.6 & 366M \\
\textbf{GPT-2}, HuggingFace & \textbf{Python} (WebText) & -- & 0.25 & 0.38 & 34.6 & 355M \\
\textbf{MultiGPT-C} & \textbf{C\#} (C\#,Python,JS,TS) & 2.01 & 0.53 & 0.66 & 74.6 & 374M \\   
\textbf{MultiGPT-C} & \textbf{Python} (C\#,Python,JS,TS) & 1.83 & 0.76 & 0.80 & 84.1 & 374M \\	
\textbf{MultiGPT-C} & \textbf{JS,TS} (C\#,Python,JS,TS) & 1.36 & 0.68 & 0.82 & 87.6 & 374M \\	
\bottomrule
\end{tabular}
\label{tab:main_results_table}	
\end{table*}
As seen, the best monolingual validation level performance in terms of edit similarity and the ROUGE-L precision and recall is achieved for Python programming language. We explain it using the notion of ``naturalness'' of source code introduced in~\cite{bigCode}. Naturalness of code has a strong connection with the fact that developers prefer to write and read code that is conventional, idiomatic, and familiar as it helps understanding and maintaining software systems, leading to code predictability. Python is also the most popular programming language, according to \textit{PopularitY of Programming Language index}\footnote{http://pypl.github.io/PYPL.html}, leading to more code adoption and reuse.

Remarkably, multilingual model achieves a comparable performance in terms of edit similarity and ROUGE-L precision, but yielding a a significantly lower ROUGE-L recall for C\# programming language. For JavaScript and TypeScript programming languages, however, all the metrics are improved with the multilingual model.




\subsection{Online Evaluation}

As we roll out IntelliCode Compose internally for performance and user experience evaluation, we have collected anonymous usage data and telemetry. Since we are not aware of any existing code completion system that offers a similar experience, selecting the correct metric to evaluate online performance was particularly challenging. The key online evaluation metrics that we are measuring are the surfacing rate (SR) and the click-through rate (CTR) over a period of time. The SR is the total number of completions displayed divided by the total number of times a completion could potentially be shown, which is after every character typed into a code document when the extension is active. The CTR is defined as the fraction of accepted completions over the total number of completions displayed. Over 150 thousands requests, we have seen a SR of 9.2\% and a CTR of 10\%, which roughly translates to suggestions being shown every 11 characters and users committing 

The SR is not only dependent on the accuracy of the model but also on the typing speed of a user and their network reliability. The low CTR can be partially attributed to the momentum in typing. As the users type, it is unlikely they will stop after every keystroke to examine suggestions. Similar to traditional code completion scenarios, users tend to overshoot for a few characters before committing a desired suggestion. Due to this momentum effect, the surfacing rate captured in our telemetry is systematically lower as compared to what a user may actually experience.

\section{Knowledge Distillation}

Knowledge distillation~\cite{hinton2015distilling} is the model compression technique in which a smaller model -- the student network -- is trained to reproduce the results of a larger model -- the teacher network. It has been shown in literature~\cite{tang2019distilling, sanh2019distilbert}, that it is possible to reach comparable performance on various tasks using distilled neural networks, resulting in models that are lighter and faster at inference time. 

Inspired by DistilBERT~\cite{sanh2019distilbert}, we scale down our pretrained transformer models by reducing the number of transformer blocks, while keeping the architecture of the transformer block and embedding layers intact. The GPT-C model size scales near-linearly with the number of transformer blocks. 

We experiment with student models having 8 and 12 transformer blocks, having our best 26 layer model serve as a teacher, initializing the student models with pretrained teacher weights and biases. Tab.~\ref{tab:distil_results_table} summarizes the distillation results for JavaScript and TypeScript programming languages, comparing it to the monolingual teacher model trained on JavaScript and TypeScript. As seen, distillation from 26 to 12 layers speeds up the inference by a factor of 2.7, incurring 6\% edit similarity loss and 5\% ROUGE-L precision loss. In a more extreme scenario, distilling a 26 layer model down to only 8 layers, we obtained a 4.5$\times$ inference speed up at a cost of 8\% edit similarity and 9\% ROUGE-L precision. 
\begin{table*}
\centering
\caption{Performance of distilled monolingual models of various sizes that are trained and evaluated on JavaScript and TypeScript programming languages, as compared to the teacher model. The inference speed is calculated using the beam search depth of $L$=25 and width of $k$=15.}
\begin{tabular}{llllllllllll} \toprule
	\textbf{Model}   &
			\multicolumn{2}{c}{\textbf{ROUGE-L}} & \textbf{Edit similarity (\%)} & \textbf{Model size} & \textbf{Inference speed} \\ \cmidrule{2-3} 
	                & Precision &  Recall \\ \midrule
    \textbf{DistilGPT-C}, tiny & 0.53 & 0.58 & 78.0 & 96M & 600ms\\
    \textbf{DistilGPT-C}, small & 0.55 & 0.65 & 79.3 & 124M & 1000ms\\	
    \textbf{GPT-C}, teacher & 0.58 & 0.72 & 84.1 & 366M & 2700ms\\	
\bottomrule
\end{tabular}
\label{tab:distil_results_table}	
\end{table*}

\section{Model Deployment}
\label{sec:deploy}

The IntelliCode Compose service is designed as two-layer service: the server-side model inference module and the client-side completion provider module. The main reason for this setup is to minimize the inference time for the best user experience. As we deploy the model on a cloud-based server, we have control over the hardware setup and can guarantee resource availability.

The server-side module is deployed as a containerized web application to Azure Kubernetes Service\footnote{https://azure.microsoft.com/en-us/services/kubernetes-service/} and listens on a HTTPS endpoint. It processes completion requests and returns the model output. It is implemented in Python and executes model inference using PyTorch and ONNX runtime\footnote{https://github.com/microsoft/onnxruntime}. We employ several graph-level model optimizations, including constant folding, and operator fusion for layer normalization and GELU sub-graphs.

The client-side completion provider is a Visual Studio Code extension implemented in TypeScript. The completion provider monitors user inputs and is responsible for communicating with the web service as well as post-processing model outputs as described in section \ref{sec:client-postprocess}.




\section{Related Work}

This work is related to a large set of literature in the area of NLP, and NLU, as well as deep learning and particularly transformers. We refer the interested reader to the numerous publications about transformer models~\cite{gpt2, bert, xlnet, roberta}, and focus on code completion for the remainder of this section. 

Numerous intelligent code completion systems for both statically and dynamically typed languages have been proposed~\cite{raychev2014code,proksch2015intelligent,bruch2009learning,asad2014methodCC}. Best Matching Neighbor (BMN) and statistical language models such as $n$-grams, as well as RNN-based approaches leveraging sequential nature of the source code have been particularly effective at creating such systems. 

Among the models that have found practical applications in IDEs are that of~\cite{svyatkovskiy2019Pythia} -- for method and API completion based on a neural language model and ASTs. The approach described in~\cite{svyatkovskiy2020fast} reformulates code completion as a task of learning to rank the valid completion suggestions computed from static analyses in order to improve computational speed and memory efficiency, effectively superseding~\cite{svyatkovskiy2019Pythia}. The code completion system based on~\cite{svyatkovskiy2019Pythia} is deployed as part of \textit{IntelliCode}~\cite{intellicode} extension in Visual Studio Code IDE, and~\cite{bruch2009learning} -- snippet matching based on frequency models and BMN -- has been deployed as part of Eclipse Code Recommenders~\cite{eclipseCodeRecommenders, eclipseSnipMatch}. 

Closest to our work is probably \textit{Tabnine}~\cite{tabnine}, which uses GPT-2 to serve ranked lists of code sequence suggestions. However, this tool does not attempt to complete longer sequences of 20--30 characters long, up to a whole line of code, and we are not aware of any currently deployed tool that does so.

\section{Conclusions}

We have introduced and deployed a general-purpose AI-powered code completion system called IntelliCode Compose, capable of generating code sequences of arbitrary token types, including local variables, methods or APIs, arguments, as well as language keywords, and delimiters. IntelliCode Compose serves as a universal programming language modeling tool, effectively generating syntactically correct code in multiple programming languages, capable of completing a whole-line of code in a couple of key strokes. 

IntelliCode Compose is built around the GPT-C -- a multi-layer generative pretrained transformer model for code, which is a variant of the GPT-2 trained from scratch on source code data. Our best model yields an average edit similarity of $86.7\%$ and perplexity of $1.82$ for Python programming language.

We have documented and overcome several practical challenges of training transformer neural networks on HPC clusters, model deployment in the cloud, and client-side caching to meet the edit-time code completion inference speed requirement of at most $100$ ms per call. We have also thoroughly studied and documented the multilingual modeling approaches on a dataset consisting of four programming languages.

In the future, we are planning to extend IntelliCode Compose capabilities by focusing on completion personalization, and fine-tuning on custom user code. Besides code completion, we plan to apply large-scale unsupervised language model pretraining on source code to tackle several other automated software engineering tasks, including automatic program repair, and code search.
\bibliographystyle{unsrt}  
\balance
\bibliography{references} 

\end{document}